\documentclass{article}
\usepackage{waspaa19,amsmath,graphicx,url,times}
\usepackage{tikz}
\usepackage{tkz-graph}

\usepackage{booktabs}       % professional-quality tables
\usepackage{color}

\title{Continual Learning of New Sound Classes using Generative Replay}

\name{Zhepei Wang$^\sharp$, 
      Cem Subakan$^\flat$,
      Efthymios Tzinis$^\sharp$,
      Paris Smaragdis${^\sharp} {^\natural}$,
      Laurent Charlin$^{\flat\diamondsuit}$ \thanks{This work is supported by NSF grant \#1453104.}
}
\address{$^\sharp$ University of Illinois at Urbana-Champaign, 
	Department of Computer Science \\       
         $^\flat$ Mila--Quebec Artificial Intelligence Institute\\
         $^\diamondsuit$ HEC Montr\'eal, Canada CIFAR AI Chair \\
         $^\natural$ Adobe Research\\       
}

\begin{document}

\ninept
\maketitle

\begin{sloppy}

\begin{abstract}
%   In this paper we propose the idea of using generative replay for continual learning with sound event classification, a setting in which we learn a model sequentially with a few classes at each stage without full access to sound classes seen before. Learning based models tend to suffer catastrophic forgetting, the degradation in performance for tasks where the training data is no longer available. We seek to remedy the issue of catastrophic forgetting with a system that consists of a classifier and a generator, both continuously learned. The model is evaluated on recordings of generic, environmental sound classes. We demonstrate that our proposed system can efficiently retrieve information from fresh, unseen sound classes while retaining the knowledge from past, unavailable data.
% In this paper,  we explore the continual learning setting for soundclassification.  Continual learning is the learning setting where weincrementally train a model on a sequence of datasets.  Althoughbeing a practically important setting, naively updating a model withnew data results in forgetting of earlier training, which is referredas  “catastrophic  forgetting”.   We  propose  a  generative  modelingscheme  for  audio  signals  to  generate  rehearsal  data  to  alleviatecatastrophic forgetting. We demonstrate on ESC-10 dataset that bycontinually training a genetive model, we are able to match the per-formance of storing 20 percent of the training dataset in the replaybuffer of a rehearsal method, by using a generator with model sizeless than the 4 percent of the training dataset.
Continual learning consists in incrementally training a model on a sequence of datasets and testing on the union of all datasets. In this paper, we examine continual learning for the problem of sound classification, in which we wish to refine already trained models to learn new sound classes. In practice one does not want to maintain all past training data and retrain from scratch, but naively updating a model with new data(sets) results in a degradation of already learned tasks, which is referred to as ``catastrophic forgetting.'' We develop a \emph{generative replay} procedure for generating training audio spectrogram data, in place of keeping older training datasets. We show that by incrementally refining a classifier with generative replay a generator that is 4\% of the size of all previous training data matches the performance of refining the classifier keeping 20\% of all previous training data. We thus conclude that we can extend a trained sound classifier to learn new classes without having to keep previously used datasets.
\end{abstract}


\section{Introduction}
\label{sec:intro}

Standard supervised machine learning setup posits that the full training dataset is available to the model at once. 
%Although there exist applications where this learning setup is useful in practice, this is somewhat different from how humans exhibit learning.
This is a simplistic assumption. In the wild, the training data may arrive in (non-iid) batches and new classes may appear throughout the learning process. This is typical of human learning where new concepts (classes) are learned throughout life. 

%\emph{Continual learning} is the machine learning setup where we try to mimic the sequential learning process that humans exhibit
\emph{Continual learning} proposes a more realistic sequential learning paradigm composed of training episodes \cite{schlimmer1986case, sutton1993online, ring1997child}. 
At each episode, the model is only trained on data from a single new task and does not have access to data from earlier tasks. Continual learning is also useful for devices with constrained access to data (either due to storage limitations, or privacy constraints). In such cases classifiers need to be continually trained to learn new classes while minimizing storage. This limits the amount of possible retraining on previous tasks.

Continual learning is particularly challenging for neural networks because of \emph{catastrophic forgetting}: at each episode the network will ``forget'' the knowledge it has learned in earlier tasks \cite{Thrun1995LifelongRL,french}. %LC2: Add references for catastrophic forgetting (ZW: added)
While a flurry of methods have been recently proposed for continual learning \cite{LopezPaz2017GradientEM, rebuffi2017icarl, Shin2017, li2018learning, parisi2018continual}, much work remains before continual learning becomes a practical technique. 
%naive training in a continual learning setup typically results in forgetting the knowledge acquired on earlier tasks. This phenomenon is referred to as the \emph{catastrophic forgetting} \cite{french} in the literature.

In this paper, we explore a continual learning setup for training a classifier on environmental sound classes. This is a challenging task because it necessitates learning a classifier on time-series data, as opposed to typical applications in the continual learning literature that focus on static data (e.g., images) \cite{rebuffi2017icarl}. %In addition to being an analog to the way in which human beings learn to classify sounds, the continual learning setup is also useful in practical applications where continual learning of new sound classes needs to be performed by devices with constrained access to data (either due to storage limitations, or privacy constraints). Such classifiers need to be continually trained to learn new sound classes while minimizing storage and retraining on data used previously.
%In this paper we design such a system. 

To alleviate catastrophic forgetting, we utilize the generative replay technique \cite{Shin2017}, which provides very competitive continually-learned classifiers. A generator is trained simultaneously with the classifier. For each task, the generator is used to simulate earlier-task examples for the classifier.  Further, we propose a convolutional autoencoder architecture to embed time-series data, and we make use of the two-step learning framework introduced in \cite{subakan2018} to learn the generative model to replay earlier tasks.

We experiment with the ESC-10 (Environmental Sound Classification) dataset \cite{esc50}. Namely, we compare our proposed generative replay based method with \emph{rehearsal} which consists in storing a fixed percentage of the data associated with earlier tasks to combat forgetting. Stored data is used as training data in each of the subsequent episodes. This method has been shown to be a very strong baseline \cite{contlearn_scenarios}. We show that by using a generative model with size approximately equal to 4\% of the whole training set, we are able to match the classification accuracy obtained with a rehearsal method which stores 20\% of the training dataset. 

\section{Methodology for Continual Learning}
\label{sec:cl}

\newcommand{\dat}{\mathcal D}
\newcommand{\gen}{\mathcal G}
\newcommand{\gencost}{\text{gencost}}
\newcommand{\ccost}{\text{cost}}
\newcommand{\Klat}{K}
\newcommand{\Lobs}{L}
\newcommand{\Mgmm}{M}

\subsection{Definition of Continual Learning}
% LC: add a reference for CL. (cem: done)
In continual learning \cite{schlimmer1986case, sutton1993online, ring1997child}, the goal is to train a model on a sequence of datasets $\{\dat_1, \dat_2, \dots, \dat_T\}$, where each dataset corresponds to a (new) \textit{task}. According to the standard continual learning setup, when training the model for task $t$, the data of past tasks and future tasks are not available. That is, when training for task $t$, we are only allowed to use the dataset $\dat_t$. The objective is to learn a single model which is able to predict well on data from all tasks $1, \dots, T$, despite training in a sequential manner. %As we alluded in the introduction, 
This is challenging in neural network models as training on the current task without incorporating data from earlier tasks typically results in forgetting the existing knowledge. This phenomenon is referred to as \emph{Catastrophic Forgetting} \cite{Thrun1995LifelongRL,french}.
Namely, when training for task $t$, the model forgets the knowledge related to tasks with index $<t$, if no measures are taken to mitigate forgetting. In the following two subsections, we describe two strategies to combat catastrophic forgetting.\footnote{ % LC: You may need to explain this in a few more sentences. (cem: I worked on the definition of forward transfer, I guess you were referring to that ? )
In addition to avoiding catastrophic forgetting, another goal of continual learning is to improve/speed-up learning on future and past tasks. This is referred to as \emph{forward transfer} and \emph{positive backward transfer} \cite{LopezPaz2017GradientEM}. This is a very interesting research direction for continual learning, but in this paper we focus more on combating catastrophic forgetting.}

\newcommand{\cem}[1]{\textbf{cem : #1}}

\subsection{Naive Rehearsal} 
        \label{sec:reh}
        %LC: In the first sentence why no briefly explain that the buffer is used to provide additional training data (D_{<t}) when training on D_t? (cem: I added the second sentence.) 
        A simple method to combat catastrophic forgetting is to keep a buffer of random samples to remember the past tasks. The buffer contains examples from earlier tasks to reinforce the knowledge from earlier tasks, when training on the current task. This method is referred to as naive rehearsal or simply \emph{rehearsal}, as we do in the rest of this paper. Although simple, this method is surprisingly effective, and has been shown to perform very comparable to state-of-the-art continual learning methods on various standard continual learning experiments \cite{contlearn_scenarios}. For this reason we use rehearsal as a baseline method.  When training for $t$, we keep a buffer $\mathcal M = \bigcup_{k=1}^{t-1} \mathcal M_k$, where $\mathcal M_k$ contains randomly selected examples from  task $k$, such that $k \leq t-1$. The cost function associated with rehearsal is: 
        	\begin{align}
				\mathcal L^t_\text{naivereplay} = \frac{1}{|\mathcal D_t|}\sum_{(x, y) \in \mathcal D_t} &\ccost(y, f_\theta(x)) + \notag \\
				& {\color{black} \sum_{k=1}^{t-1} \frac{1}{|\mathcal M_k|} \sum_{(x', y') \in \mathcal M_k } \ccost(y', f_\theta(x')),}
			\end{align}
    where the first term accounts for the loss on the current task (current loss), and the second term accounts for the rehearsal loss. The input features are denoted with $x$, the target values are denoted with $y$, the continually trained classifier is denoted with $f_\theta(.)$, and $\ccost(.)$ denotes a classification loss, which is typically chosen as the cross-entropy loss. In Figure \ref{fig:basic_rh} we illustrate the schematics of the loss function. 
        
        \newcommand{\blockcolor}{black}
        \begin{figure}[h!]
			\begin{center}
			\begin{tikzpicture}[node distance=2cm,auto,>=latex']
				\node (2) [draw=black, solid, node distance=1cm]{$f^t$};
				\node (1) [left of=2]{$\mathcal X^t$};
				\node (3) [right of=2]{$\hat{\mathcal Y}^t$};
				\node (4) [right of=3]{current loss};

				\node (a) [below of=1, node distance=1cm]{{\color{\blockcolor}$\mathcal X^{1:t-1}_\text{buffer}$}}; 
				\node (b) [below of=3, node distance=1cm]{{\color{\blockcolor}$\hat{\mathcal Y}^{1:t-1}$}};
				\node (c) [right of=b]{{\color{\blockcolor}rh. loss}};

				\draw[->] (1) edge (2);
				\draw[->] (2) edge (3);
				\draw[->] (3) edge node {$\mathcal Y^t$} (4);
				\draw[\blockcolor,->] (b) edge node {\color{\blockcolor}{$\mathcal Y^{1:t-1}_\text{buffer}$}} (c);

				\draw[\blockcolor,->] (a) edge (2);
				\draw[\blockcolor,->] (2) edge (b);
			\end{tikzpicture} 
			\end{center}
			\caption{The diagram for the loss computation in the naive rehearsal method at task $t$. We separately compute two losses for the current tasks, and a rehearsal (rh.) loss on the stored buffer. }
			\label{fig:basic_rh}
        \end{figure}
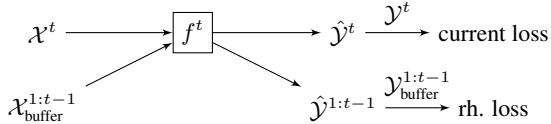
   %LC: below add a reference. (cem: I removed the decision boundary claim I had earlier) 
    Even though rehearsal combats forgetting, it requires the storage of data in form of a rehearsal buffer. In the next section, we introduce another method which mitigates forgetting by continually learning a generative model, which does not require storage of past data items. 

\subsection{Generative Replay}
    An effective alternative to rehearsal is generative replay \cite{Shin2017}. This method continually trains a generative model in addition to the classifier to \emph{replay} the data from earlier tasks. By the virtue of having a generative model, in lieu of storing examples from earlier tasks, we generate data, and use this generated data to avoid forgetting (by using it as training data).
    The cost function for continual classifier training is therefore written as follows:

\begin{align}
	    {\mathcal L}^t_\text{genreplay} = \sum_{(x, y) \in \dat_t} \ccost( f_t(x), y ) + 
	    {\color{black} \sum_{x_g \in \dat_g} \ccost(f_t(x_g), f_{t-1}(x_g))}, \label{eq:genreplay}
    \end{align}
    where, the first term is the loss associated with the current task, and the second term is the loss associated with the rehearsal, where $\dat_g$ is data simulated from the generative model model after being done with training it until task $t-1$, which is used to rehearse the datasets $\{ \dat_1, \dots, \dat_{t-1}\}$. The schematic illustration of this loss function is shown in Figure \ref{fig:genreplay_classifier}. Similarly, the generator $G^t$ is trained by using the examples from the current dataset $\dat_t$ and the simulated examples from the generator $G^{t-1}$:
    
    \begin{align} 
	    {\mathcal L}^t_{\text{gen}} = \sum_{x \in \dat_t} \gencost( x ) + {\color{black} \sum_{x_g \in \dat_g} \gencost( x_g )}, \label{eq:contgen}
    \end{align}
where again the loss function is composed of the current loss term (the first term) and the rehearsal loss (the second term). We illustrate the workflow of the method in Figure \ref{fig:genreplay_gen}. 

    \label{sec:genrep}
    \begin{figure} [h!]
        \centering
        		\begin{tikzpicture}[node distance=2cm,auto,>=latex']
			        \node [draw=black, dashed] (a) {$G^{t-1}$};
				\node (b) [right of=a] {$\mathcal X^{1:{t-1}}_\text{replay}$};
				\node (c) [draw=black, solid, right of=b] {$G^t$};
				\node (d) [right of=c] {replay loss};
				\node (e) [below of=b, node distance=1cm] {$\mathcal X^t$};
				\node (f) [below of=d, node distance=1cm] {current loss};

				\draw[->] (a) edge (b);
				\draw[->] (b) edge (c);
				\draw[->] (c) edge (d);
				\draw[->] (e) edge (c);
				\draw[->] (c) edge (f);
			\end{tikzpicture} 

        \caption{Diagram for continually training a generative model using generative replay: At task $t$, the data is \emph{replayed} from the generative model $G^{t-1}$, and its likelihood is evaluated on the generative model $G^t$ that we currently train. The dashed blocks means that the parameters are frozen, and not being updated, and solid blocks mean that the block parameters are optimized. }
        \label{fig:genreplay_gen}
    \end{figure}
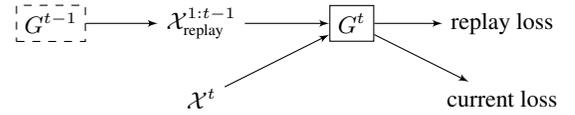
    
    \begin{figure}[h!]
        \centering
        			\begin{tikzpicture}[node distance=2cm,auto,>=latex']
			 
			    \node (a) [] {$\mathcal X^{1:t-1}_\text{replay}$};
			    
				\node (b) [draw=black, dashed, right of=a] {$f^{t-1}$};
				\node (c) [right of=b] {$\mathcal Y^{1:t-1}_\text{target}$};
				\node (d) [right of=c] {replay loss};
				\node (e) [draw=black, solid, below of=b, node distance=1cm] {$f^t$};
				\node (f) [below of=c, node distance=1cm] {$\hat{\mathcal Y}^{1:t-1}$};
				\node [draw=black, dashed, below of=a, node distance=1cm] (h) {$G^{t-1}$};
				\node (2) [below of=e, node distance=1cm]{};
				\node (1) [left of=2]{$\mathcal X^t$};
				\node (3) [right of=2]{$\hat{\mathcal Y}^t$};
				\node (4) [right of=3]{current loss};

                \draw[->] (h) edge (a);
				\draw[->] (a) edge (b);
				\draw[->] (b) edge (c);
				\draw[->] (c) edge (d);
				\draw[->] (a) edge (e);
				\draw[->] (e) edge (f); 
				\draw[->] (f) edge (d);

				\draw[->] (1) edge (e);
				\draw[->] (e) edge (3);
				\draw[->] (3) edge node {$\mathcal Y^t$} (4);
			\end{tikzpicture} 

        \caption{Training the classifier using the generative replay at task $t$: The data for the earlier tasks is generated from $G^{t-1}$. The outputs of the current classifier $f^t$ and the earlier classifier $f_{t-1}$ are matched to compute a replay loss.}
        \label{fig:genreplay_classifier}
    \end{figure}
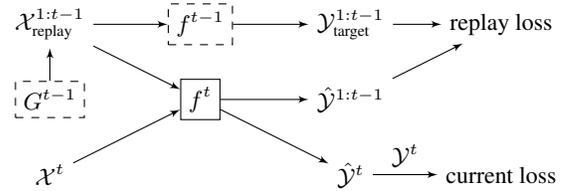
    Note that in our application the generator $G$ generates of spectra segments as our goal to classify segments of audio data. Next, we describe the details of the architecture of the generator $G$. 

\subsection{The Generative Model Architecture}
\label{sec:2-step-lrn}

In this paper, we use maximum-likelihood based generative modeling as opposed to Generative Adversarial Networks (GANs) \cite{NIPS2014_5423} as the former is significantly easier to train \cite{Lucic2018}. 
%LC: above add a reference. (cem: done)

In our generative models, we use a convolutional autoencoder to compute embeddings for spectrogram sequences. The architecture of our autoencoder is shown in Figure \ref{fig:AE_arc}, which consists of using convolutional layers across the time axis to model the temporal structure, and then reducing and increasing the feature dimensionality using fully connected layers.
%LC: The dimensionality of what? (cem : done)
After learning the embeddings $h$, we learn the generative model by fitting a Gaussian mixture model (GMM) on the latent embeddings, as described in the 2-step learning method in \cite{subakan2018}. Advantages of using GMMs in the latent space is advocated by multiple papers in the literature \cite{hoffman2016elbosurgery, subakan2018, Jiang2016, Dilokthanakul2016, Tomczak2017}. 
% LC: State explictely what the 2-state approach is. (cem : done)
In our experiments we have observed that separating the learning of parameters of the prior distribution on the latent variables from the learning of autoencoder resulted in the accurate learning of the generative model (which we refer to as 2-step training). We have observed that the joint training of GMM and the autoencoder often resulted in slightly worse results than that of the 2-step learning approach, and therefore we have chosen to use the 2-step training rather than jointly training the prior and the autoencoder. We also compare the proposed generative modeling scheme with VAEs with standard Gaussian prior \cite{Kingma2013}, and observe that the proposed generative modeling scheme yields much superior generations, which results in better classification.

\begin{figure}[h!]
    \centering
    \includegraphics[width=0.44\textwidth]{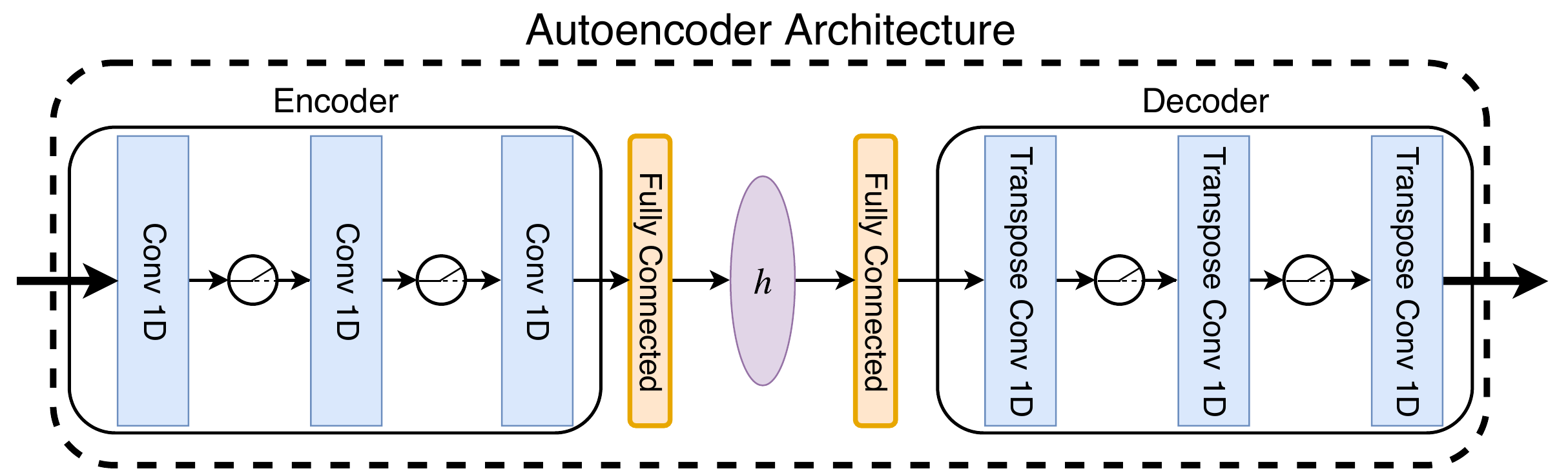}
    \caption{The autoencoder architecture used to model the spectra. The convolutional encoder maps the spectra into latent space $h$, which is then transformed by the decoder into reconstructed representation. We apply ReLU after each of the first two convolutional layers in both the encoder and the decoder.}
    \label{fig:AE_arc}
\end{figure}

\section{Experimental Setup}
\label{sec:exp}

In this section we introduce our continual learning setup for audio classification. The experiments simulate scenarios where the model incrementally learns new sound classes without having full access to the previously-encountered sound
%LC: seen -> heard? :)  (cem: or, encountered ? )
classes. The model observes ten sound classes in a sequence of five tasks, where in each task two new classes are presented. This is similar to similar to the setup in~\cite{closed-loop-gan}.
% LC: May be worth justifying this particular setup. (cem: Zhepei? -> ZW: reference added) 

\subsection{Data}
\label{ssec:data}
We select the publicly available ESC-10 \cite{esc50} dataset for our experiments. The ESC-10 dataset consists of $400$ five-seconds recordings sampled at 44kHz of acoustic events from $10$ classes, namely: \textit{chainsaw}, \textit{clock ticking}, \textit{crackling fire}, \textit{crying baby}, \textit{dog barking}, \textit{helicopter}, \textit{rain}, \textit{rooster}, \textit{seawaves}, and \textit{sneezing}.

For each recording we extract a Time-Frequency (TF) spectrogram representation using a $2048$ samples window and a $512$ samples hop size. Next, we compute the square root of the mel-scaled spectrogram using $128$ mel-features for each spectrogram. We further segment our data to snippets that correspond to $\approx 220$ ms so that each input data sample has a size of $128 \times 16$. We ignore low-energy spectra whose Frobenius norm is less than 1e-4. Finally, we normalize each spectrogram by the maximum energy from each mel-spectrogram so that each value lies in $[0,1]$. Our initial experimental results demonstrate that normalized mel-spectrograms are more discriminative under the chosen classifier architecture and can be easily reconstructed from the generator. In total there are $9500$ mel-spectrograms that we further split into training, validation and test set with a ratio of $7:2:1$. 

To setup the experiment in the setting of continual learning, we partition the dataset into five subsets/tasks where all classes are mutually exclusive. We group the classes based on the their label indices so the two sound classes from the same group are more similar to each other compared to classes from the other groups. 

\subsection{Generative Replay Setup}
\label{ssec:gr_setup}
We next discuss the setup for generative replay including the architecture of the classifier and the generator.

\subsubsection{Classifier Architecture}
\label{sssec:clf_arch}
The classifier contains two 1-D convolutional layers with $64$ and $128$ filters, respectively, one average pooling layer and two fully-connected layers with $50$ and $10$ hidden nodes each. For the convolutional layers, we use a filter of length $3$ and perform same-padding to the input. We use a rectified linear unit (ReLU) as a nonlinearity after each convolutional layer and the first fully-connected layer. The output of the second fully-connected layer is passed into a softmax layer for a $10$-class classification.

\subsubsection{Generator Architecture}
\label{sssec:gen_arch}
We experiment with both the autoencoder and the variational autoencoder architectures as the generator. The encoder consists of three 1-D convolutional layers followed by a fully-connected layer with $50$ hidden units. Each of the convolutional layers uses $128$ filters of lengths $6$, $4$, and $3$ and strides of $1$, $2$, and $2$, respectively. The decoder consists of three 1-D transposed convolutional layers, each with $128$ filters of length $4$, $4$, and $7$ and stride of $2$, $2$, and $1$, respectively. We do not perform zero-padding and we apply ReLU after each one of the first two convolutional layers in both the encoder and the decoder as shown in Figure \ref{fig:AE_arc}. The variational autoencoder architecture contains an additional linear layer on top of the convolutional encoder with $50$ dimensions with a reparameterization trick for being able to sample from the latent space.

\subsection{Rehearsal Setup}
\label{ssec:rhs}
We compare the proposed generative replay mechanism with rehearsal based methods. We set up the rehearsal data by storing $p\%$ of the training data at each task into a buffer. This buffer is available to the models throughout all tasks. In our setting, the size of the buffer increases linearly with the number of tasks. We adjust the percentage of the rehearsal data such that the the data from each task have equal probability to be drawn. The parameter of the percentage $p$ of the rehearsal data lies in $p \in \{5, 10, 20, 100\}$.

\subsection{Training Setup}
\label{ssec:training}
For all experiments, we optimize our models using Adam\cite{adam}. The batch size is set to $100$, and there are $10$ - $15$ batches per epoch for each task. To train the classifier, we use an initial learning rate equal to 5e-4 and we train it for 300 epochs by minimizing the cross-entropy loss for each task. Moreover, in order to train the generator, we use an initial learning rate of 1e-3 and train it for 1700 epochs for each task. The autoencoder loss is the binary cross-entropy for each time-frequency bin between the original spectrogram and the reconstruction. The loss for the variational autoencoder is the sum of binary cross-entropy and KL-Divergence between the modeled distribution and unit Gaussian.

\section{Results and Discussions}
\label{sec:res}
We report the performance of various replay strategies under the sound classification setup. For each experiment we report the performance obtained by the models using five different permutations of the order of tasks. In each task, we report the mean accuracy on the test set, which contains all sound classes that the model has seen up until the current task.
\begin{figure}[htb!]
    \centering
    \includegraphics[width=0.48\textwidth]{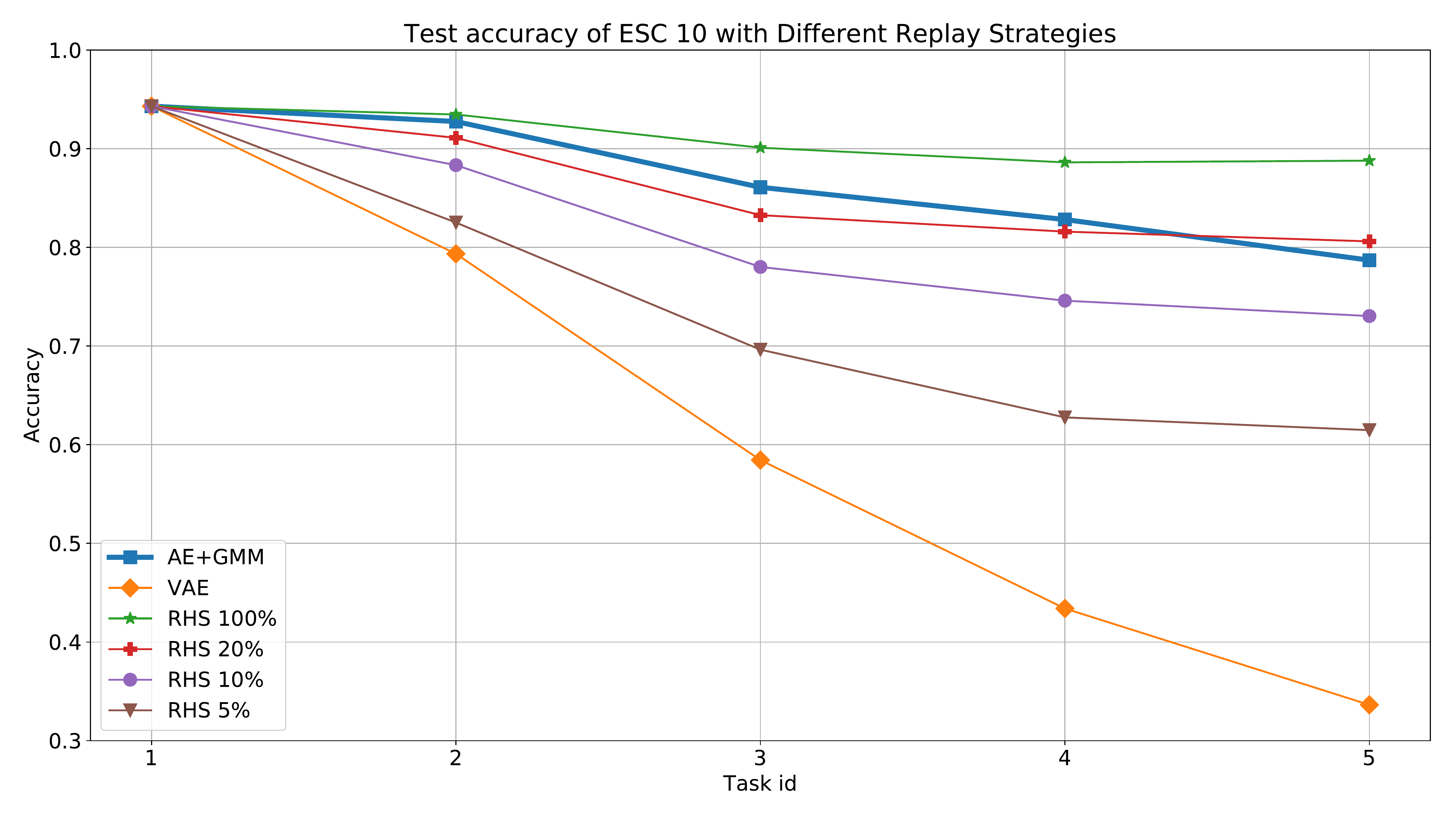}
    \caption{Test accuracy on the ESC-10 dataset using generative replay and rehearsal methods with various buffer sizes. The x-axis denotes the task index and the y-axis refers to the accuracy of the model's prediction. Each point represents the mean of the accuracy after five runs using different permutations for the tasks.} %LC: If space discuss resuults. (cem: I guess in section 4.3, we have good discussion?)
    \label{fig:acc}
\end{figure}
\subsection{Overall Results}
Figure~\ref{fig:acc} shows the test accuracy of different generative replay strategies and rehearsal methods for various buffer sizes. ``AE+GMM'' refers to the proposed generative replay setting with an autoencoder and a Gaussian mixture learned in two steps as described in Section~\ref{sec:2-step-lrn}. ``VAE'' corresponds to the variational autoencoder mentioned in Section~\ref{sssec:gen_arch}. ``RHS X\%'' denotes a rehearsal based method with X\% training data stored into the buffer. ``RHS $100$\%'' is used as an upper-bound estimation of the performance of any replay strategy since it corresponds to the ideal case where all the training data is available in all future stages.

Overall, RHS $100$\% has the highest mean accuracy and it fits the expectation as an upper-bound estimation of any replay strategy. The performance of rehearsal methods increases as the proportion of the data stored in buffer increases. We also notice that for all methods the variance of the test accuracy tends to decrease as the number of tasks increases. Initially, the variance is large because a random binary classification task might deviate too much in terms of difficulty from another. However, towards the end, the models have seen all sound classes regardless of the permutation, and therefore the mean accuracy tends to stabilize.

\subsection{Comparison Between AE+GMM and VAE}
AE+GMM significantly outperforms VAE as a replay strategy. The mean accuracy of AE+GMM is similar to RHS 20\%, while VAE performs significantly worse than 5\%. We analyze such notable difference by looking at the samples generated by both models as illustrated in Figure~\ref{fig:gen_samples}. We show three examples from the training set and the respective generations using AE+GMM and VAE. Note that VAE smooths out the temporal structure of the generated mel-spectrograms and lacks diversity between classes. On the other hand, AE+GMM generates mel-spectrograms with much more diverse temporal structure, exhibiting much closer resemblance to the examples from the training set.
%LC: aligned? Do you mean that the structure of the AE+GMM resembles the one of the training samples? (cem : I proposed the above sentence) 
\begin{figure}[htb!]
\centering
    \includegraphics[width=0.48\textwidth]{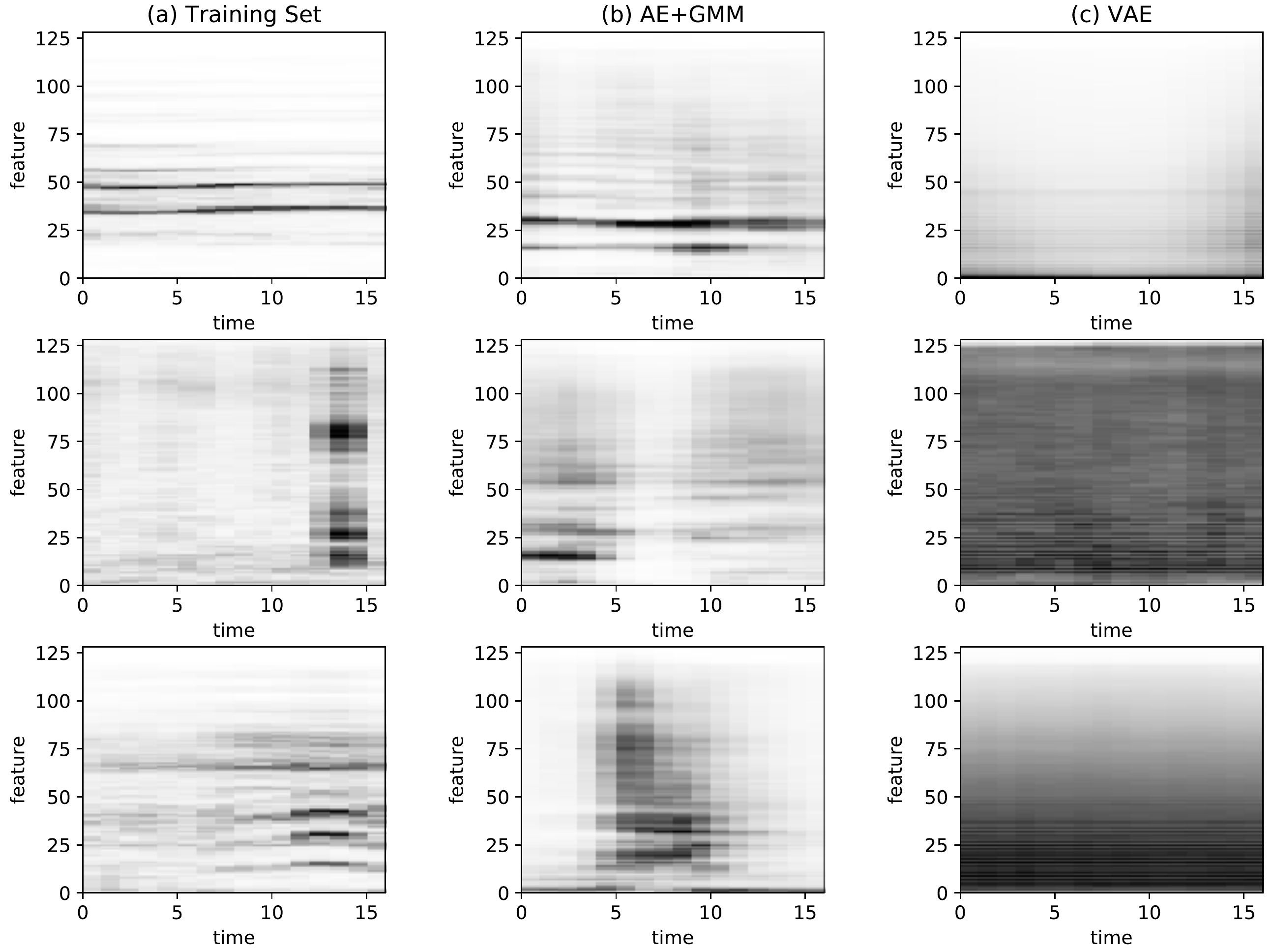}
\caption{Mel-spectrograms from the training set (column a), generated by AE-GMM (column b) and generated by a VAE (column c). Notice how the VAE generated data do not reproduce salient class features, whereas the proposed AE+GMM generator does so better.
}
\label{fig:gen_samples}
\end{figure}
\subsection{Comparison Between AE+GMM and Rehearsal Based Methods}
We observe that AE+GMM performs significantly better than rehearsal schemes with buffer proportion $p = 5\%, 10\%$. The accuracy of AE+GMM is almost identical to RHS 20\% at the last task and marginally higher in all previous tasks. The total number of trainable parameters in AE+GMM is less than $480,000$. The size of the network is equivalent to $\frac{480000 / (128\times 16) }{9500 \times 0.7} \approx 3.5\%$ of the training data. In other words, using a generator whose size is less than $4$\% of the training data, we are capable of reaching the accuracy comparable to storing $20$\% of the data. The result demonstrates the effectiveness of AE+GMM generative replay strategy when limited storage space is available.

\section{Conclusion}
\label{sec:conc}

We showed that generative replay is an effective continual learning method for audio classification tasks. 
Using a generative model whose size is less than $4$\% of the size of the training data, we obtain a test accuracy comparable to a buffer-based rehearsal scheme which needs to store $20$\% of all used training data. These results highlight the potential of using generative models instead of keeping previously seen training data when there are storage constraints. We see these aspects being crucial to sound recognition systems for which keeping prior training data is prohibitive, but often need (to learn) to perform new tasks on the fly.
%cem : above, ... for which storing training data examples is prohibitive, .... 

%In summary, this work is an initial attempt of using generative replay to incrementally learn a classifier on time-series data.
%LC: Future work is a bit generic. (cem: I think Zhepei commented out the last sentence) 
%Our approach can also extend to powerful future systems that can handle sound classification with more classes, more tasks, and sound with more dynamic structures.

% -------------------------------------------------------------------------
% Either list references using the bibliography style file IEEEtran.bst
\bibliographystyle{IEEEtran}
\bibliography{refs19}
%
% or list them by yourself
% \begin{thebibliography}{9}
% 
% \bibitem{waspaa19web}
%   \url{http://www.waspaa.com}.
%
% \bibitem{IEEEPDFSpec}
%   {PDF} specification for {IEEE} {X}plore$^{\textregistered}$,
%   \url{http://www.ieee.org/portal/cms_docs/pubs/confstandards/pdfs/IEEE-PDF-SpecV401.pdf}.
%
% \bibitem{PDFOpenSourceTools}
%   Creating high resolution {PDF} files for book production with 
%   open source tools, 
%   \url{http://www.grassbook.org/neteler/highres_pdf.html}.
%
% \bibitem{eWilliams1999}
% E. Williams, \emph{Fourier Acoustics: Sound Radiation and Nearfield Acoustic
%   Holography}. London, UK: Academic Press, 1999.
% 
% \bibitem{ieeecopyright}
%   \url{http://www.ieee.org/web/publications/rights/copyrightmain.html}.
%
% \bibitem{cJones2003}
% C. Jones, A. Smith, and E. Roberts, ``A sample paper in conference
%   proceedings,'' in \emph{Proc. IEEE ICASSP}, vol. II, 2003, pp. 803--806.
% 
% \bibitem{aSmith2000}
% A. Smith, C. Jones, and E. Roberts, ``A sample paper in journals,'' 
%   \emph{IEEE Trans. Signal Process.}, vol. 62, pp. 291--294, Jan. 2000.
% 
% \end{thebibliography}

\end{sloppy}
\end{document}